\documentclass[11pt]{article}

\usepackage[final]{EMNLP2022}

\usepackage{times}
\usepackage{latexsym}

\usepackage[T1]{fontenc}

\usepackage[utf8]{inputenc}

\usepackage{microtype}

\usepackage{float}
\usepackage{inconsolata}
\usepackage{multirow}
\usepackage{xcolor}
\usepackage{devanagari}
\usepackage{graphicx} 
\usepackage{placeins}

\title{SIT at MixMT 2022: \\
Fluent Translation Built on Giant Pre-trained Models}

 \author{Abdul Rafae Khan, Hrishikesh Kanade, Girish Amar Budhrani, \\ {\bf Preet Jhanglani}, and {\bf Jia Xu} \\
        Stevens Institute of Technology\\
         \{akhan4, hkanade, gbudhran, pjhangl1, jxu70\}@stevens.edu}

\begin{document}
\maketitle

\begin{abstract}
This paper describes the Stevens Institute of Technology's submission for the WMT 2022 Shared Task: Code-mixed Machine Translation (MixMT). The task consisted of two subtasks, subtask $1$ Hindi/English to Hinglish and subtask $2$ Hinglish to English translation. Our findings lie in the improvements made through the use of large pre-trained multilingual NMT models and in-domain datasets, as well as back-translation and ensemble techniques. The translation output is automatically evaluated against the reference translations using ROUGE-L and WER. Our system achieves the $1^{st}$ position on subtask $2$ according to ROUGE-L, WER, and human evaluation, $1^{st}$ position on subtask $1$ according to WER and human evaluation, and $3^{rd}$ position on subtask $1$ with respect to ROUGE-L metric.
\end{abstract}

\section{Introduction}
\label{sec:intro}
Code-mixing (or code-switching) is the phenomenon when another language like Hindi is interleaved with English words in the same sentence. This code-mixed language is mostly used in social media text and is colloquially referred to as Hinglish. Despite Hindi being the fourth most widely spoken language in the world \cite{ethnologue}, research in Hinglish translation has been a relatively unexplored task.


A major challenge in creating a translation system for code-mixed text is the limited amount of parallel data \cite{ranathunga2021neural}. Typical methods use standard back-translation techniques \cite{sennrich2015improving} for generating synthetic parallel data for training. Massive multilingual neural machine translation (NMT) models have recently been shown to improve the translation performances for low-resource and even zero-shot settings. We propose using such large multilingual NMT models for our code-mixed translation tasks.


Previous work has only used smaller multilingual architectures \cite{gautam2021comet}. We use pre-trained multilingual models trained in up to 200 language directions. We finetune these models for the Hindi to Hinglish and Hinglish to English tasks. One major challenge when using these massive models is the GPU memory constraint. Another issue is the ratio of English and Hinglish words interleaved for each translation output. We use multiple state-of-the-art GPUs with model parallelization to overcome the memory issue. For the amount of English in the outputs, we tune the model parameters including learning rate, dropout, and the number of epochs to get the optimal translations.

Along with these pre-trained multilingual NMT models, we also use additional in-domain data, back-translation to generate additional parallel data, and using multi-run ensemble to improve the final performance. All these methods gave us an improvement of $+24.4$ BLEU for Hindi to Hinglish translation (subtask 1) and $+28.1$ BLEU points for Hinglish to English translation (subtask 2) compared to using only the organizer provided data and the baseline experiment.

In this paper, we discuss our submission for the WMT 2022 MixMT shared task. We participate in both the subtasks and our submission system includes the following:
\begin{itemize}
    \item Tune very large pre-trained multilingual NMT models and finetune on in-domain datasets;
    \item Back-translation to create synthetic data for in-domain monolingual data;
    \item Multi-run ensemble to combine models trained on various datasets;
    \item Tune model parameters to enhance model performance.
\end{itemize}

\section{Related Work}
\label{sec:related}
\paragraph{Multilingual Neural Machine Translation (MNMT)}
Word and subword-level tokenizations are widely used in natural language processing, including NMT/MNMT. ~\citet{morishita2018improving} propose to incorporate hierarchical subword features to improve neural machine translation. Massively multilingual NMT models are proposed by ~\citet{aharoni2019massively} and ~\citet{arivazhagan2019massively}. They are trained on a large number of language pairs and show a strong and positive impact on low-resource languages. However, these models tend to have representation bottlenecks ~\cite{dabre2020survey}, due to the large vocabulary size and the large diversity of training languages.
Two MNMT systems ~\cite{tan2019multilingual,xiong2021contrastive} are proposed to solve this problem by modifying the model architectures, adding special constraints on training, or designing more complicated preprocessing methods.~\citet{xiong2021contrastive} adopt the contrastive learning scheme in many-to-many MNMT.~\citet{tan2019multilingual} propose a distillation-based approach to boost the accuracy of MNMT systems. However, these word/subword-based models still need complex preprocessing steps such as data augmentation or special model architecture design. 


\paragraph{Code-mixed NMT}

The majority of research for code-mixed translation focuses on generating additional data using back-translation methods. \citet{winata2019code} used the sequence to sequence models to generate such data and \citet{garg2018code} used a recurrent neural network along with a sequence generative adversarial network. \citet{pratapa2018language} generated linguistically-motivated sequences. Additionally, there have been several code-mixed workshops \cite{bhat2017joining,aguilar2018proceedings} to advance the field of code-mixed data.

\paragraph{Hindi or Hinglish NMT}
Researchers have worked on machine translation from Hindi to English \cite{laskar2019neural,goyal2019ltrc}, however, there has been far less work for Hinglish. A major issue is the lack of parallel Hinglish-English data. Additional parallel data generated by back-translation is used to improve the performance \cite{gautam2021comet,jawahar2021exploring}. 
The CALCS'21 competition \cite{solorio2021proceedings} had a shared task for English to Hinglish for movie review data. 

\section{Background}

\subsection{Task Description}

The WMT 2022 CodeMix MT task consists of two subtasks. Subtask 1 is to use Hindi or English as input and automatically translate it into Hinglish. Subtask 2 is to input a Hinglish text and translate it into English. Participation in both subtasks was compulsory for the competition. We use Hindi only as the source for subtask 1.

\subsection{Neural Machine Translation}
The Neural Machine Translation (NMT) task uses a neural network-based model to translate a sequence of tokens from one human language to another. More formally, given a sequence of tokens in source language $x = \{x_1,x_2,\cdots ,x_n\}$, the model outputs another sequence of tokens in target language $y = \{y_1,y_2,\cdots ,y_m\}$. The input sequence $x$ is encoded into the latent representation by a neural network-based encoder module, and these representations are decoded by the neural network-based decoder module. We train transformer-based encoder-decoder models \cite{vaswani2017attention} to translate the data. These models use a self-attention mechanism in their architectures to boost performance.

\subsection{Multilingual NMT (MNMT)}
Initial NMT systems were only capable of handling two languages. However, lately, there has been a focus on NMT models which can handle input from more than two languages \cite{dong2015multi,firat2016multi,johnson2017google}. Such models, commonly called Multilingual NMT (MNMT) models, have shown improvement in low-resource or zero-shot Neural Machine Translation settings. Instead of translating a sequence of tokens in source language $x$ to another sequence in target language $y$, the MNMT system uses multiple sources and target languages. 

There are two main approaches: (1) use a separate encoder and decoder for each of the source and target languages \cite{gu2018universal}, and (2) use a single encoder/decoder which shares the parameters across the different languages \cite{johnson2017google}.

The issue with the first approach is that it requires a much larger memory due to multiple encoders and decoders \cite{vazquez2018multilingual}. The second approach is much more memory efficient due to parameter sharing \cite{arivazhagan2019massively}. 

Training a model using the second approach can be done by adding a language tag to the source and target sequence. Specifically, when the decoding starts, an initial target language tag is given as input, which forces the model to output in that specific language.

\section{Methods}



For the initial set of experiments, we use the baseline transformer model \cite{vaswani2017attention}. For all the other experiments, we use pre-trained multilingual NMT models and fine-tuned them for the specific datasets. We can divide these into three groups based on the number of parameters. (1) smaller models including mBART-50 \cite{tang2020multilingual} and Facebook M2M-100 medium model \cite{fan2021beyond} (M2M-100), (2) the medium models include the Facebook NLLB-200 \cite{costa2022no} (NLLB-200) and Google mT5 XL \cite{xue-etal-2021-mt5} (mT5-XL), and (3) for large model we use the Google mT5 XXL model \cite{xue-etal-2021-mt5} (mT5-XXL). The parameter count for each of the models and the training time per epoch for baseline datasets are mentioned in Table \ref{tab:model_stats}.

For both subtasks, we use Hindi as the source language tag and English as the target language tag. 


\subsection{Pre-trained Models}

To train the transformer, mBART-50, and M2M-100 models, we use the Fairseq toolkit \cite{ott2019fairseq}, and the larger NLLB-200, mT5-XL, and mT5-XXL models use the Huggingface toolkit \cite{wolf2019huggingface}. Table \ref{tab:model_stats} lists the parameter count for each pre-trained multilingual model.


\begin{table}[H]
    \centering
    \resizebox{.25\textwidth}{!}{
        \begin{tabular}{c|c}
            Model & Params \\\hline
            mBART-50 & 611M \\
            M2M-100 & 1.2B \\
            NLLB-200 & 3.3B \\
            mT5-XL & 3.7B  \\
            mT5-XXL & 13B \\
        \end{tabular}
    }
    \caption{Parameter count for each pre-trained multilingual model.}
    \label{tab:model_stats}
    \vspace*{-1em}
\end{table}
\subsection{Data Augmentation}
\label{subsec:data_aug}
We use three different ways to add additional in-domain data for training our models. 

\paragraph{Additional in-domain data} 
We use additional in-domain parallel data and add it to the training data for accuracy improvement. Since our focus is on Hindi for subtask 1 and Hinglish for subtask 2, we tried to look for data from additional domains with Hindi or Hinglish as the source. We use Kaggle Hi-En \cite{HinditoHinglish2020} and MUSE Hi-En dictionary \cite{lample2017unsupervised} for subtask $1$. For subtask $2$, we use Kaggle Hg-En data \cite{Louis_Tom}, CMU movie reviews data \cite{zhou-etal-2018-dataset}, and CALCS'21 Hg-En dataset \cite{solorio2021proceedings}. We also use selected WMT'14 News Hi-En sentences \cite{bojar2014hindencorp} and the MTNT Fr-En and Ja-En data \cite{michel2018mtnt}. Table \ref{tab:data_stats1} all lists these datasets.



\paragraph{Back-translation}
A common technique used to increase the data size for low-resource languages is to use in-domain monolingual data and generate synthetic translations using a reverse translation system \cite{sennrich2015improving}. We use google translate for back-translation. We translate samples from the English side of Tatoeba Spanish to the English dataset \cite{Tatoeba} and Sentiment140 dataset \cite{go2009twitter} into Hinglish and use the synthetic translations as additional bilingual data.

\subsection{Ensemble}
We use a multi-run ensemble \cite{koehn2020neural} to combine multiple model's best checkpoints to boost the final performance. We average the probability distribution over the vocabulary for all the models to generate a final probability distribution and use that to predict the target sequence.


\begin{table}[!ht]
    \centering
    \resizebox{.50\textwidth}{!}{
        \begin{tabular}{c|c|c|c}
            Dataset & Sentences & $V_R$ & $V$ \\\hline
            HinGE Hi-Hg & 2.3K & 103K & 19K\\
            PHINC Hg-En & 13K & 302K & 55K \\\hline
            HinGE Hg-En & 11K & 109K & 22K\\
            Kaggle Hi-En & 11K & 220K & 31K\\
            Kaggle En-Hg & 1.8K & 98K & 17K\\
            MUSE Hi-En & 30K & 29K & 24K \\
            CMU Reviews Hg-En & 8K & 180K & 24K \\
            CALCS'21 Hg-En & 8K & 182K & 23K \\
            Back-translation Hg-En & 8.5K & 48K & 7K \\\hline
            WMT'14 Hi-En & 15K & 181K & 21K \\
            MTNT Fr-En & 10K & 16K & 14K  \\
            MTNT Ja-En & 3.5K & 120k & 8K\\
        \end{tabular}
    }
    \caption{Datasets provided by the organizers and additional in-domain and out-of-domain datasets used for subtask 1 and 2. $V_R$ is the number of running words and $V$ is the vocabulary size.}
    \label{tab:data_stats1}
    \vspace*{-1.2em}
\end{table}

\section{Datasets}

The competition provided one dataset for each of the subtasks, HinGE Hi-Hg \cite{srivastava-singh-2021-hinge} for subtask 1 and PHINC Hg-En \cite{srivastava-singh-2020-phinc} for subtask 2. The competition also provided the validation data. In addition to these, we also use additional in-domain and out-of-domain datasets. 


Due to a large overlap of English and Hinglish vocabulary, we use Hindi-English (Hi-En) and Hindi-Hinglish (Hi-Hg) datasets for subtask 1. For subtask 2, we use various Hinglish-English datasets. All the competition provided datasets, the additional in-domain datasets, and the additional out-of-domain datasets used for both the subtasks are listed in Table \ref{tab:data_stats1}. As HinGE En-Hg has multiple Hinglish translations for a single English sentence. We duplicated the English to increase the size of the data. For the WMT'14 Hi-En dataset, we selected the closest $15$K sentences, selected using cosine similarity with source-side validation data.

%
%
%

To preprocess the data, we tokenize using the Moses tokenizer \cite{MOSESDECODER} or the model-specific tokenizer provided by Huggingface. Additionally, we apply either Byte pair encoding (BPE) \cite{sennrich2015neural} for the baseline transformer model and sentence piece \cite{sentencepiece} for all other models including mBART-50, M2M-100, NLLB-200, mT5-XL and mT5-XXL to split words into subwords tokens.


\section{Experiments}

This section describes the experimental details, including the toolkits, the parameter settings for the model training and decoding, and the results.

\subsection{Tools \& Hardware}
\label{sec:tools}



For the Models mentioned in Section \ref{subsec:data_aug}, we train the smaller models on 32GB NVIDIA Tesla V100 GPUs, and the medium and larger models require multiple 80GB NVIDIA A100 GPUs.  We use a total of 4 V100 GPUs and 16 A100 GPUs. Due to GPU memory usage (see Section \ref{sec:intro}), we parallelized the training of the medium and larger models using the DeepSpeed package \cite{rasley2020deepspeed}. 


\subsection{Training Details}


As an NMT baseline, we use the baseline transformer model \cite{vaswani2017attention} provided by the Fairseq toolkit. The model has half number of attention heads and the feed-forward network dimension compared to the Transformer (base) model in \citet{vaswani2017attention}.  The rest of the network architecture is the same. We train this model from scratch by adding additional datasets and finally tuning it on the validation data.

We use the Fairseq toolkit for training the baseline transformer from scratch and for finetuning the mBART-50 and M2M-100 models. For finetuning NLLB-200, mT5-XL, and mT5-XXL models, we use the Huggingface toolkit. For the pre-trained multilingual models, we use the Hindi language encoder and English language decoder for finetuning and decoding.

As shown in Table \ref{tab:xfmr_exp}, we finetune the models with the listed datasets for each subtask. We initially fine-tune these models on ID 4 dataset mentioned in Table \ref{tab:xfmr_exp}. Finally, we further finetune the models on the validation datasets provided by the organizers.

\begin{table}[!ht]
    \centering
    \resizebox{.4\textwidth}{!}{
        \begin{tabular}{c|c|c}
            Model &  \multicolumn{2}{c}{Train time/epoch} \\\cline{2-3}
             & Subtask 1 & Subtask 2 \\\hline
            mBART-50 & 2 mins & 14 mins \\
            M2M-100 & 8 mins & 33 mins \\
            NLLB-200 & 16 mins & 1.5 hrs\\
            mT5-XL & 20 mins & 15 hrs \\
            mT5-XXL & 5.5 hour & 24 hrs \\
        \end{tabular}
    }
    \caption{Per epoch training time for each of the models. The training time is for ID 4 datasets in Table \ref{tab:xfmr_exp}.}
    \label{tab:train_stats}
    \vspace*{-1.5em}
\end{table}

\begin{table*}[!ht]
    \begin{minipage}{.50\linewidth}
      \centering
      \resizebox{.75\textwidth}{!}{
        \begin{tabular}{c|c|c}
              ID& Datasets  & Hi-Hg \\\hline
               1& HinGE & 1.2\\
               2& [1]+Kaggle & 6.4\\
               3& [2]+WMT'14 News & 10.3\\
               4& [3]+Facebook MUSE & 10.5\\\hline
               5& [4]+val tune & 11.6
            \end{tabular}
            }
    \end{minipage}%
    \begin{minipage}{.50\linewidth}
      \centering
      \resizebox{.75\textwidth}{!}{
            \begin{tabular}{c|c|c}
                   ID & Datasets  & Hg-En \\\hline
                    1 & PHINC & 4.5\\
                    2 & [1]+HinGE & 5.1\\
                    3 & [2]+CALCS'21  & 5.2\\
                    4 & [3]+Back-translation & 8.5 \\\hline
                    5 & [4]+val tune & 8.7
                \end{tabular}
            }
    \end{minipage} 
    \caption{Adding in-domain datasets. Baseline: Transformer~\cite{vaswani2017attention}. Evaluation critierion: BLEU[\%].  add citation of the datasets. Training from scratch without pre-trained models. `+val tune' is further finetuning on validation data. All the results are evaluated on the competition's test data. }
    \label{tab:xfmr_exp}
    \vspace*{-0.7em}
\end{table*}

\begin{table*}[!ht]
    \centering
    \resizebox{.80\textwidth}{!}{
        \begin{tabular}{c|c|c||c|c}
            \multirow{2}{*}{Pretrained Multilingual Model} & \multicolumn{2}{c||}{subtask 1} & \multicolumn{2}{|c}{subtask 2}\\\cline{2-5}
             & baseline & +val tune & baseline & +val tune\\\hline
            mBART-50 & 16.9 & - & 18.3 & - \\
            M2M-100 & 18.9 & - & 23.8 & - \\
            NLLB-200 & 11.5 & - & 23.8 & 30.3 \\
            mT5-XL & 18.8 & \textbf{25.6} & 24.0 & 31.7 \\
            mT5-XXL & 18.5 & 24.0 & 24.9 & \textbf{32.6} \\
        \end{tabular}
    }
    \caption{Initialization with pre-trained models. BLEU scores (\%) for subtask 1 and 2. `baseline' experiment is finetuning the pre-trained model on the ID 4 datasets in Table \ref{tab:xfmr_exp}. `+val tune' is further finetuning on validation data. All the results are evaluated on the competition's test data. \textbf{bold} results are the final submission.
    }
    \label{tab:results}
    \vspace*{-1em}
\end{table*}

\paragraph{Hyper-parameter settings} 
We train the Transformer model from scratch and finetune all the multilingual pre-trained models. We train Transformer, mBART-50, and M2M-100 models for $10$ epochs on the ID 4 datasets and $5$ epochs on the validation dataset. We finetune the larger models listed in Table \ref{tab:train_stats}, for a maximum of 3 epochs before tuning on the validation for 7 epochs for subtask 1 and 4 epochs for subtask 2, respectively. We set the Adam betas to 0.9 and 0.98 for all the models and tuned the learning rates between $1e^{-5}$ and $9e^{-5}$. We opt for higher learning rates for the initial epochs and use lower learning rates for the remaining epochs. Finetuning with a high learning rate for fewer epochs is particularly helpful as larger models take much more time per epoch, even with the larger GPU memory. We also experiment with tuning the dropout between 0.1 and 0.15, and we get the best performance with the dropout rate set to 0.1. The batch size is limited to smaller values due to memory constraints. We set the batch size to 10 or 20 for larger models and 40 or 50 for medium-sized or smaller models. 



\paragraph{Decoding parameters}

For the decoding step for both tasks, we set English as the target language tag for all the models. We tune the beam size, and the optimal beam size is $17$ for both subtasks on the validation set. We limit the maximum sentence length to $128$ only for the medium and larger models like NLLB-200, mT5-XL, and mT5-XXL. Finally, we detokenize the translation output as a post-processing step~\cite{MOSESDECODER}.

\subsection{Additional Experiments}
\label{sec:data_aug}

We also perform additional experiments that are helpful but not included in the final submission due to limited time. These are the MTNT datasets and the ensemble methods. 
Firstly, we use the MTNT dataset as an additional bilingual in-domain data set containing different source languages. 
We also apply the multi-run ensemble method to combine models trained on multiple datasets together ~\cite{Koehn17}. For both tasks, we train M2M-100 models on the MTNT Fr-En data and the MTNT Ja-En data before tuning them on the baseline datasets, respectively. Additionally, we first fine-tune the WMT'14 News Hi-En data and then fine-tune the baseline data. Then we ensemble these two models with the original base model.

\section{Results}


We evaluate the models with respect to the BLEU score using \verb|sacrebleu|. Table \ref{tab:results} shows the results of the experiments for both tasks and all the models. In general, we get improvement with larger multilingual models and with validation finetuning.

Table \ref{tab:xfmr_exp} shows the results of training from scratch using the transformer model with additional in-domain datasets. We get a maximum improvement of $9.3$ for subtask 1 and $4.0$ for subtask 2 using the additional datasets. Finally, tuning on validation gave an additional boost of $+1.1$ and $+0.2$ BLEU for subtasks 1 and 2 respectively. Table \ref{tab:results} shows the results for using pre-trained multilingual models on the ID 4 datasets. We get a maximum improvement of $25.6$ and $32.6$ for subtasks 1 and 2. This is $+14.0$ and $+23.9$ BLEU points higher than the best transformer model's results in Table \ref{tab:xfmr_exp}.
%


Table \ref{tab:add_exp} shows the ensemble results of a multi-run ensemble of the three models: (1) The baseline M2M-100 model in Table \ref{tab:results}, (2) The M2M-100 model first trained on MTNT data and then on the baseline data, and (3) Training the M2M-model on MTNT data, then on WMT data, and finally on the baseline data. We get a slight decrease of $-0.3$ BLEU for subtask 1 compared to the baseline. However, for subtask 2, the performance improves by $+0.8$ BLEU points.

\begin{table}[!ht]
    \centering
    \resizebox{.46\textwidth}{!}{
        \begin{tabular}{c|c|c}
           Task & Models & BLEU \\\hline
            \multirow{2}{*}{subtask 1} & Base  & 18.9 \\
            & Base+MTNT+WMT & 18.6 \\\hline
            \multirow{2}{*}{subtask 2} & Base  & 23.8 \\
            & Base+MTNT+WMT & 24.6 \\
        \end{tabular}
    }
    \caption{Checkpoint ensemble results for subtask 2 trained on M2M-100 model evaluated on the competition's test data. The base is the baseline M2M-100 experiment. MTNT is first training on MTNT data and then tuning on the baseline. WMT tunes on MTNT, then WMT, and finally on baseline data.}
    \label{tab:add_exp}
    \vspace*{-1em}
\end{table}

\section{Analysis}
\label{sec:analysis}

We analyze the translation outputs of NLLB, mT5-XL, and mT5-XXL models. 
%
%
For subtask 1, the issues we faced were that the sentences were translated entirely to English and did not contain any Hinglish words. Some words were translated partially to Hinglish, and a portion of the words remained in the Hindi language. 
For subtask 2, the issues we faced were that the names of animal species were not translated correctly. And idioms lose their meaning in translation. 
Examples of these issues are shown in Table \ref{tab:analysis1} \& \ref{tab:analysis2}. 



\begin{table}[!ht]
    \centering
    \resizebox{.52\textwidth}{!}{
    \begin{tabular}{l|l}
        Src & {\small{\dn d\?f kF rA{}\6{\3A3w}F{}y E\387w{}k\?{}V VFm}} ... \\
        NLLB & The national cricket team in the country...\\
        mT5-XL & desh ki national cricket team... \\
        mT5-XXL & country ki national cricket team... \\
        Ref & desh ki national cricket team...\\\hline
        Src & {\small {\dn \305w\qq{h} \qq{\3FEw}mAEZt ho \7{c}kA h\4 jo ek} .{\dn cm(kAr h\4}} \\
        NLLB & It has been proven which is a miracle.\\
        mT5-XL & yah pramanit ho chuka hai jo ek miracle hai.  \\
        mT5-XXL & yah pramanit ho chuka hai jo ek {\dn cm }tkaar hai.
        \\
        Ref & yah pramanit ho chuka hai jo miracle hai. \\
    \end{tabular}
    }
    \caption{Examples of errors for subtask 1.}
    \label{tab:analysis1}
    \vspace{-1em}
\end{table}

\begin{table}[!ht]
    \centering
    \resizebox{.5\textwidth}{!}{
    \begin{tabular}{l|l}
        Src & lol...gayi bhains paani mein... \\
        NLLB & lol... went bhains in water... \\
        mT5-XL & lol... animals went in water... \\
        mT5-XXL & Lol... Goat got in the water... \\
        Ref & lol.. buffalo went in the water...\\\hline
        Src & ye video dekh kar to khoon khaul gya \\
        NLLB & After seeing this video, blood came out. \\
        mT5-XL & seeing this video, my blood bleed. \\
        mT5-XXL & Blood boiled after watching this video. \\
        Ref & By watching this video, blood boiled. \\
    \end{tabular}
    }
    \caption{Examples of errors for subtask 2.}
    \label{tab:analysis2}
    \vspace*{-1em}
\end{table}




\section{Conclusion}

This paper describes our submitted translation system for the WMT 2022 shared task MixMT competition. We train five different multilingual NMT models including mBART-50, M2M-100, NLLB-200, mT5-XL, and mT5-XXL, for both subtasks. We finetune on in-domain datasets including the validation data and significantly enhance our translation quality from $1.2$ to $25.6$ and $4.5$ to $32.6$ for subtasks 1 and 2 respectively. Additionally, we also apply data-augmentation techniques including back-translation, tuning on in-domain data, and checkpoint ensemble. Our system got the 1$^{st}$ position in subtask 2 for both ROUGE-L and WER metrics, the 1$^{st}$ position in subtask 1 for WER, and 3$^{rd}$ position in subtask 1 for ROUGE-L.

\section*{Acknowledgments}
We appreciate the National Science Foundation (NSF) Award No. 1747728 and NSF CRAFT Award, Grant No. 22001 to fund this research. We are also thankful for the support of the Google Cloud Research Program. We especially thank 
Xuting Tang, Yu Yu, and Mengjiao Zhang to help editing the paper. 

\bibliography{anthology,custom}
\bibliographystyle{acl_natbib}



\end{document}